\documentclass[10pt,conference]{IEEEtran}
\IEEEoverridecommandlockouts
\usepackage{cite}
\usepackage{amsmath,amssymb,amsfonts}
\usepackage{algorithmic}
\usepackage{graphicx}
\usepackage{textcomp}
\usepackage{xcolor}
\usepackage{subfig}
\usepackage{booktabs}
\usepackage{import}
\usepackage{color}
\usepackage[capitalise]{cleveref}
\usepackage{multirow}
\usepackage{array}
\usepackage[font=footnotesize]{caption}
\usepackage[inline]{enumitem}
\usepackage{makecell}
\usepackage{siunitx}
\usepackage{caption}
\usepackage{soul}
\usepackage{amsfonts}
\usepackage{booktabs}
\usepackage{subcaption}
\usepackage[top=0.75in, bottom=1.05in, left=0.75in, right=0.75in]{geometry}
\def\BibTeX{{\rm B\kern-.05em{\sc i\kern-.025em b}\kern-.08em
T\kern-.1667em\lower.7ex\hbox{E}\kern-.125emX}}

\usepackage{tikz}
\newcommand\copyrighttext{%
  \footnotesize \textcopyright 2026 IEEE. Personal use of this material is permitted. Permission from IEEE must be obtained for all other uses, in any current or future media, including reprinting/republishing this material for advertising or promotional purposes, creating new collective works, for resale or redistribution to servers or lists, or reuse of any copyrighted component of this work in other works.}
\newcommand\copyrightnotice{%
\begin{tikzpicture}[remember picture,overlay]
\node[anchor=south,yshift=10pt] at (current page.south) 
  {\fbox{\parbox{\dimexpr\textwidth-\fboxsep-\fboxrule\relax}{\copyrighttext}}};
\end{tikzpicture}%
}

\begin{document}
\bstctlcite{IEEEexample:BSTcontrol}
%

\title{Deciphering Region-Level Signatures from Latency Measurements in LEO Satellite Internet}

\author{\IEEEauthorblockN{
    Xiang Shi,~Yifei~Zhang,~and~Peng~Hu
\IEEEauthorblockA{
    Advanced Network and Embedded Systems Lab (AEL)\\
    Dept. of Electrical and Computer Engineering, University of Manitoba, Winnipeg, Canada}
    Email: peng.hu@umanitoba.ca
}
\\
\thanks{We acknowledge the support provided by the Government of Canada, Natural Sciences and Engineering Research Council of Canada (NSERC), [funding reference number RGPIN-2022-03364], and Research Manitoba.}
}

\maketitle
\copyrightnotice
\begin{abstract}
Low-Earth orbit (LEO) satellite Internet has become an indispensable infrastructure that provide growing coverage for global users. Despite extensive measurement efforts, the principles underlying region-level performance characteristics remain insufficiently understood, limiting the ability to identify region-specific latency signatures under dynamic network conditions. In this paper, we formulate the problem of region-level latency characterization using Starlink round-trip time (RTT) measurements from the public LENS dataset. We then propose a hierarchical analytical framework that transforms raw RTT sequences into multi-scale statistical features for cross-region comparison. Using data from five geographically representative regions, we demonstrate that latency differences are strongly associated with deployment factors, particularly infrastructure availability and Starlink dish-to-Point-of-Presence distance. Mutual information analysis identifies minimum RTT as the most discriminative feature, which is further supported by XGBoost-based feature importance. The proposed model well achieves 83\% accuracy on short-term data. However, its performance degrades over longer periods, indicating limited temporal generalization and motivating the need for adaptive models and feature representations for long-term performance in the future.

\end{abstract}


%
\IEEEpeerreviewmaketitle

\vspace{-5pt}
\section{Introduction}



Low-Earth orbit (LEO) satellite Internet has become an important non-terrestrial Internet infrastructure. With growing number of global users, it is projected to expand Internet coverage to remote, unserved, and underserved areas. However, despite its widespread availability and growing network measurement data, we have not yet established principles for understanding the performance of satellite Internet at an area and regional level. For example, it is evident that the performance of Starlink varies significantly across different locations \cite{Hu2024}. Therefore, it is essential to develop a mechanism for understanding the region-specific differences that can adapt to unique characteristics of satellite networks and their deployment. 


LEO satellite network measurements and analysis can help improve network operational efficiency 
and fault diagnostics \cite{Amin2023}. Recent studies have improved our understanding of Starlink performance through large-scale measurement campaigns, delay characterization, and latency prediction~\cite{zhao2024lens, mult, Dem, one-way, cas,chen2025minimum}. Some of the work has shown that round-trip time (RTT) varies across geographic regions and is closely related to infrastructure proximity and routing conditions. While other studies further reveal fine-grained temporal dynamics in Starlink latency, including short-timescale variation and recurring structural patterns. In sum, these efforts provide valuable empirical insights which mainly focus on measurement, characterization, or prediction from a system-wide perspective. In contrast, less attention has been paid to interpretable region-level characterization based directly on fine-grained RTT traces. Although raw RTT measurements contain rich temporal information, they are too high-frequency and noisy for direct cross-region comparison. This motivates the need for compact statistical representations that can summarize local latency behavior while preserving the characteristics most relevant to regional differences. Accordingly, this paper asks the following question: Do Starlink RTT traces exhibit region-dependent statistical signatures, and which features are most informative for characterizing them? To answer these questions, the main contributions of this paper are summarized as follows:

\begin{itemize}
\item We formulate the problem of region-level latency characterization in LEO satellite Internet using Starlink RTT measurements from the LENS dataset; 
\item We propose a hierarchical analytical framework that converts raw RTT sequences into multi-scale statistical features for cross-region comparison;
\item We demonstrate regional latency differences are associated with geographic deployment factors and identify the most informative features for region discrimination.
\end{itemize}

To the best of the authors’ knowledge, prior work has not explored the RTT measurements for region-level identification. The rest of the paper is structured as follows: Section II discusses the related work. Section III introduced the problem formulation. Section IV discusses the proposed analytical framework. Section V presents evaluation results, followed by the conclusive remarks and future work made in Section VI.

\section{Related Work}
Network measurement is an important area of Internet research. It has traditionally focused on the terrestrial networks and has more recently extended to the non-terrestrial networks (NTN), in particular LEO satellite Internet. Although Internet measurement platforms are well established, measurement of LEO satellite networks usually requires operator-specific hardware at user terminals. Such hardware enables fine-grained observations of LEO satellites involved in the data transmission and can capture their network's inherent mobility and time-varying dynamics. 

The research community has focused on the dataset generation for LEO satellite Internet. LEOScope from the University of Surrey is the first testbed that enables the measurements of Starlink LEO satellite Internet at the global scale. Arising from this work, the LENS dataset~\cite{zhao2024lens} is one of LEO satellite network measurement datasets, containing fine-grained measurement observations. It provides geographically distributed Starlink dish measurements with timestamped RTT observations and point-of-presence (PoP) related deployment information, enabling empirical analysis of latency behavior across regions. Moreover, its hierarchical organization by region and time supports both fine-grained temporal analysis and systematic regional comparison, making it well suited for latency characterization tasks in Starlink networks. WetLinks~\cite{wetlink} offers a large-scale longitudinal Starlink dataset collected from two European vantage points over six months, comprising RTT, throughput, packet loss, traceroutes, and co-located weather measurements, which makes it particularly useful for studying long-term performance trends and weather effects. In parallel, Casparsen \textit{et al.}~\cite{cas} analyze fine-grained end-to-end Starlink latency using 500 Hz active uplink (UL), downlink (DL), and RTT measurements, revealing a repeatable 15-second structure that can support short-term latency prediction and service-availability estimation. These datasets collectively form an important empirical basis for studying spatial and temporal performance variability in Starlink networks.


Recent studies have advanced the empirical and statistical understanding of Starlink and LEO satellite latency. Mohan \textit{et al.}~\cite{mult} conduct a large-scale multi-source study of Starlink performance using M-Lab, RIPE Atlas, and controlled terminal experiments, showing that latency varies substantially across regions and is closely related to the deployment of ground stations and PoPs. Izhikevich \textit{et al.}~\cite{Dem} propose HitchHiking, a scalable methodology for measuring LEO satellite networks through Internet-exposed services, and show that Starlink latency is strongly influenced by nearby PoP availability and inter-satellite-link routing. Going beyond large-scale RTT observation, Garcia \textit{et al.}~\cite{one-way} provide a detailed characterization of Starlink one-way delay using high-frequency probe measurements, revealing minor diurnal variation, clear uplink/downlink asymmetry, and strong latency inflation during periodic 15-second reconfiguration events. Similarly, Casparsen \textit{et al.}~\cite{cas} develop a statistical framework for end-to-end latency characterization and prediction, showing that Starlink latency exhibits a repeatable 15-second structure with identifiable boundary spikes and stable intra-period behavior. Taken together, these works demonstrate that latency in LEO satellite Internet is shaped by both geographic factors and fine-grained temporal dynamics, while also motivating more interpretable frameworks for systematic regional latency characterization. 

\section{Problem Formulation}

This section formalizes the research problem addressed in this paper. We consider the RTT measurements collected from multiple Starlink regions. Each region produces a high-frequency latency time series, and our objective is to determine whether different regions exhibit distinguishable latency patterns.
For each region, a one-hour RTT measurement data is represented as a timestamped sequence:

\[
H = \left\{\left(\tau_i, r_i\right)\right\}_{i=1}^{N}
\]

where $\tau_i$ denotes the timestamp of the $i$-th sample, $r_i$ is the corresponding RTT value and $N$ is the total number of observations. 
Although the raw RTT sequence contains detailed latency information, it is too fine-grained for direct region-level comparison. Therefore, the problem is to transform raw RTT observations into compact, interpretable statistical representations that retain essential temporal and distributional characteristics while enabling comparisons across regions.
Specifically, this paper addresses three objectives: (i) to characterize region-dependent latency patterns using statistical features derived from fine-grained RTT observations; (ii) to investigate whether observable geographic factors, e.g., Starlink dish to PoP ground distance, are associated with cross-region RTT differences; and (iii) to identify the most informative features for distinguishing one region from another, thereby revealing which RTT statistics best capture region-dependent latency signatures. These objectives establish the foundation for the hierarchical analytical framework and feature-based regional comparison developed in the following sections.

\section{Proposed Analytical Framework}

\subsection{Framework Overview}
To address the problem formulated in Section III, we propose a hierarchical analytical framework for region-level latency characterization. Since raw RTT measurements are high-frequency and too fine-grained for direct cross-region comparison, the main idea is to transform them into compact and interpretable statistical representations at multiple temporal scales. Specifically, the framework consists of three stages: 1) segmenting the raw RTT sequence into non-overlapping one second (1-s) intervals and extracting primary statistical features, 2) aggregating these second-level features over 60 seconds (60-s) sliding windows to generate more stable secondary representations, and 3) using the resulting feature space for regional comparison, discriminative analysis, and downstream classification. In this way, the proposed framework preserves short-term latency dynamics while providing robust descriptors for cross-region analysis.

\subsection{Primary Feature Extraction}
To capture short-term latency dynamics, we first partition the raw RTT sequence into non-overlapping 1-s segments. This design reduces the granularity of the original measurements while preserving local temporal behavior within each second. For each segment, we extract a set of primary statistical features, including measures of central tendency, variability, extrema, and tail latency. These features provide a compact summary of the local RTT distribution and serve as the basis for the subsequent window-level aggregation. 

Formally, let the $k$-th 1-s segment be defined as
\[
S_k = \{(\tau_i, r_i) \in H \mid \lfloor \tau_i \rfloor = k\},
\]
where $k$ denotes the $k$-th second of the measurement period and $n_k = |S_k|$ is the total number of RTT samples in the segment. Since the original RTT measurements are collected every 10 ms, each segment contains approximately 100 samples under normal conditions.

For each segment $S_k$, we compute a set of descriptive statistics to summarize the local RTT distribution. The sample mean and standard deviation are defined as
\[
\mu_k = \frac{1}{n_k}\sum_{(\tau_i,r_i)\in S_k} r_i,
\]
\[
\sigma_k = \sqrt{\frac{1}{n_k-1}\sum_{(\tau_i,r_i)\in S_k}(r_i-\mu_k)^2}.
\]

We further compute the median, minimum, and maximum RTT values as
$
m_k = \mathrm{Median}\{r_i : (\tau_i,r_i)\in S_k\},
$, 
$
\min_k = \mathrm{Min}\{r_i : (\tau_i,r_i)\in S_k\},
$, and 
$
\max_k = \mathrm{Max}\{r_i : (\tau_i,r_i)\in S_k\}.
$

To capture tail latency, we additionally calculate the empirical quantiles
\[
q_{k,\alpha} = \mathrm{Quantile}_{\alpha}\{r_i : (\tau_i,r_i)\in S_k\}, \quad \alpha \in \{0.95, 0.99\}.
\]

The resulting primary feature vector for the $k$-th segment is
\[
\theta_k = [n_k,\mu_k,\sigma_k,m_k,\min_k,\max_k,q_{k,0.95},q_{k,0.99}].
\]
These primary features jointly characterize the local RTT distribution in terms of central tendency, variability, extrema, and tail behavior.

\subsection{Secondary Feature Aggregation}
Although the primary features provide useful second-level summaries of RTT behavior, they are still too fine-grained for robust region-level comparison. Therefore, we further aggregate the primary feature vectors over sliding windows to capture longer-term statistical characteristics of latency. Specifically, we use a sliding window of 60 s with a step size of 30 seconds. For each window, we compute a set of secondary features that summarize the average behavior, variability, and percentile structure of the primary statistics within that interval. In addition, we introduce a fluctuation metric based on the average absolute difference between consecutive second-level mean RTT values, which quantifies the short-term temporal variation of latency. Through this design, each window is mapped into a compact and more stable feature representation for downstream regional comparison and discriminative analysis.

Let the number of 1-s segments within a sliding window be denoted by $n_w$. To quantify short-term fluctuation within a window, we define the average absolute difference of consecutive second-level mean RTT values as

\[
\delta_w = \frac{1}{n_w - 1}\sum_{k=1}^{n_w-1} |\mu_{k+1} - \mu_k|,
\]

where $\mu_k$ denotes the mean RTT of the $k$-th 1-s segment in the window. This metric reflects the smoothness or volatility of latency evolution over time.

Following this design, each sliding window is represented by a 14-dimensional secondary feature vector. Features 1--7 correspond to the window-level averages of the primary statistics, while the remaining features capture fluctuation, extrema, median behavior, and high-percentile characteristics. The definitions of the 14 secondary features extracted from each 60-s sliding window are summarized in Table~\ref{tab:60-sec_features}.

\begin{table}[htbp]
    \centering
    \caption{Summary of Statistical Features for Sliding Windows}
    \label{tab:60-sec_features}
    \resizebox{0.45\textwidth}{!}
    {
        \begin{tabular}{c|c|c}
            \toprule
            \textbf{Index} & \textbf{Feature Variable Name} & \textbf{Description}\\
            \hline
            F1  & \texttt{rtt\_s\_mean\_w\_mean} & Average of 1-s mean RTT\\ 
            F2  & \texttt{rtt\_s\_std\_w\_mean} & Average of 1-s std RTT\\
            F3  & \texttt{rtt\_s\_median\_w\_mean} & Average of 1-s median RTT\\
            F4  & \texttt{rtt\_s\_min\_w\_mean} & Average of 1-s min RTT\\
            F5  & \texttt{rtt\_s\_max\_w\_mean} & Average of 1-s max RTT\\
            F6  & \texttt{rtt\_s\_P95\_w\_mean} & Average of 1-s $P_{95}$ RTT\\
            F7  & \texttt{rtt\_s\_P99\_w\_mean} & Average of 1-s $P_{99}$ RTT\\
            F8  & \texttt{rtt\_s\_mean\_absdiff\_mean} & Mean absolute difference $\delta_w$\\
            F9  & \texttt{rtt\_s\_max\_w\_max} & Maximum of 1-s max RTT\\
            F10 & \texttt{rtt\_s\_min\_w\_min} & Minimum of 1-s min RTT\\
            F11 & \texttt{rtt\_s\_median\_w\_median} & Median of 1-s median RTT\\
            F12 & \texttt{rtt\_s\_std\_w\_std} & Std of 1-s std RTT\\
            F13 & \texttt{rtt\_s\_P95\_w\_P95} & $P_{95}$ of 1-s $P_{95}$ RTT\\
            F14 & \texttt{rtt\_s\_P99\_w\_P99} & $P_{99}$ of 1-s $P_{99}$ RTT\\
            \bottomrule
        \end{tabular}
    }
\end{table} 

\subsection{Analysis Pipeline}
Based on the hierarchical feature construction described above, the proposed framework supports three complementary analyses. First, the primary 1-s features are used for direct cross-region comparison of RTT statistics, such as minimum, median, and mean latency, in order to examine whether stable regional separation exists. Second, the secondary sliding-window features are used to assess the discriminative importance of different statistics through mutual information analysis. Third, the same window-level feature representation serves as the input to a downstream classifier for region identification and feature importance analysis. Therefore, the framework provides a unified pipeline that links fine-grained RTT measurements, interpretable statistical feature extraction, and region-level latency analysis.

\section{Evaluation}

\subsection{Dataset description}
In this section, we will review the LENS dataset’s data acquisition methodologies, the data we utilized in our analysis, and the rationale for our selection.

\subsubsection{Measurement dish location selection}
\begin{figure}[htbp]
  \centering
  \includegraphics[width=1\linewidth, height=5cm]{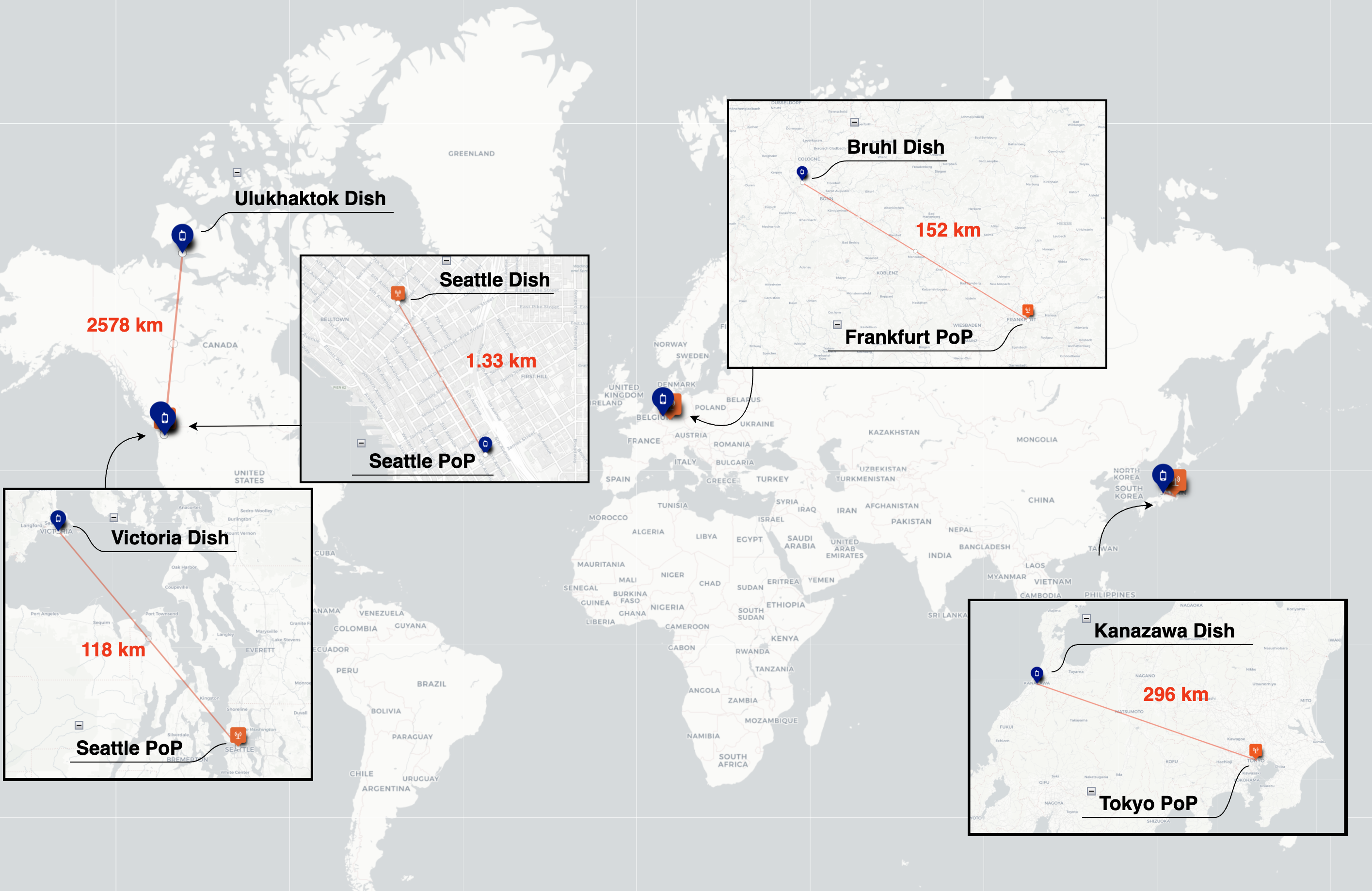}
   \caption{Starlink dish, PoP stations and dish-PoP distances.}
  \label{fig:map}
\end{figure}

From the 30 Starlink dish measurement sites in the LENS dataset, we select dish measurement data from five different regions for analysis (e.g., Victoria, Ulukhaktok, Seattle, Bruhl, and Kanazawa). Victoria, Ulukhaktok, and Seattle are in North America, while Bruhl is in Europe, and Kanazawa is in Asia. Furthermore, the LENS dataset~\cite{zhao2024lens} provides explicit coordinates between user dishes and their associated PoP stations. We can therefore calculate the ground distance between the dish and the PoP locations. We indicate the dishes and PoPs' locations on the world map (Fig. \ref{fig:map}) and their corresponding distance. There are three rationales for selecting the dishes from these five regions for our analysis. First, we can ensure precise locations for the dishes and their corresponding PoP stations. Second, we confirm that all selected dishes had active subscriptions. Finally, all these dishes contain the latest available measurement data in the LENS dataset. 

\subsubsection{LENS dataset structure}
The LENS dataset adopts a hierarchical storage structure. Data is first categorized by region. Within each regional directory, files are organized by year and month. Within these monthly directories, files are further partitioned by date. Each date directory contained hourly .txt files, each storing two features: the timestamp ($\tau$) and RTT in milliseconds. \\
A one-hour dataset for a specific region is denoted as $H$: 

\begin{equation}
\begin{aligned}
H &= (\tau_{(1)}, r_{(1)}), (\tau_{(2)}, r_{(2)}), \ldots, (\tau_{(N)}, r_{(N)}), \\
&\quad \text{such that} \quad
\tau_{(1)} \le \tau_{(2)} \le \cdots \le \tau_{(N)}.
\end{aligned}
\label{eq:hourly_data}
\end{equation}

In (\ref{eq:hourly_data}), $N$ denotes the total number of samples within the file. The LENS measurement time interval is 10 ms, so the expected values of $N$ are 360,000 (i.e., $N = 100 \times 60 \times 60$).

\subsection{Cross-region Comparison}
\begin{figure}[htbp]
    \centering
    \includegraphics[width=1\linewidth, height=6.5cm]
    {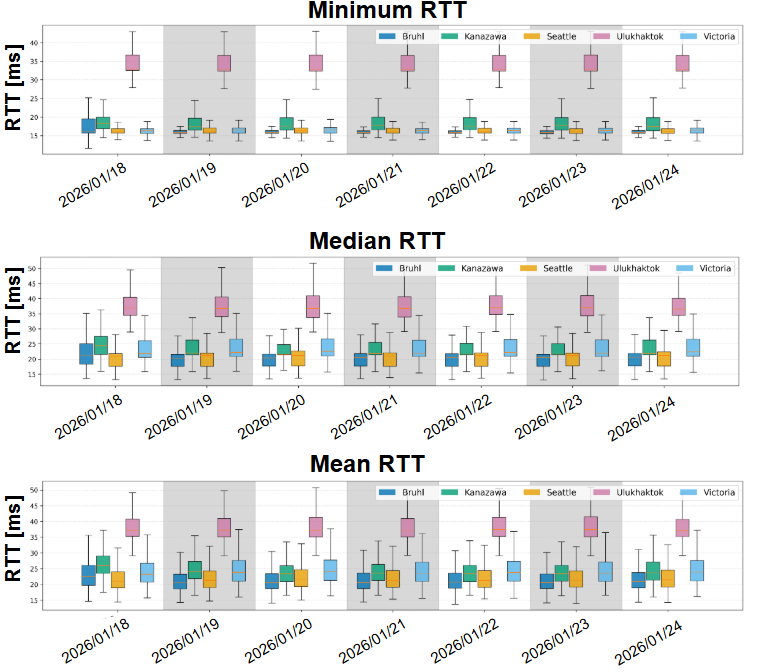}
    \caption{Statistical summary of 1-s segment feature vectors ($\theta_k$) for five different regions in seven-day (Jan 18–24, 2026).}
    \label{fig:experiment1.1}
\end{figure}

In this section, we utilize the 1-s primary features ($\theta_k$) to analyze and compare RTT distributions across the five different regions. As illustrated in Fig.~\ref{fig:experiment1.1}, the Ulukhaktok region shows a statistically significant separation from other regions, exhibiting consistently higher RTT values across all three features (e.g., minimum, median, and mean RTT). While regions like Bruhl, Seattle, and Victoria maintain a minimum RTT of approximately 15-18 ms, Ulukhaktok stays within the 35-40 ms range. The mean and median features also clearly represent that the Ulukhaktok region's RTT is significantly higher than that of other regions. 

Our findings are highly consistent with recent studies (e.g., ~\cite{bose2025investigating} and~\cite{mult}), which identify infrastructure proximity as the dominant factor impacting the network performance. Specifically, Bose \textit{et al.}\cite{bose2025investigating} points out that in regions with limited infrastructure, RTT increases significantly due to the extreme distance to the nearest PoP station. As shown in Fig.~\ref{fig:map}, the ground distance between Ulukhaktok's dish and its associated Seattle PoP station is significantly greater than that of other regions. In summary, our empirical evidence directly confirms that regions with limited infrastructure and long ground distances between the dish and the PoP station lead to increased network latency, as described in prior literature.

\subsection{Mutual Information Analysis}

In this section, we employ a nonparametric method based on k-nearest neighbors (KNN) entropy estimation mutual information (MI)~\cite{ross2014mutual} methodology to evaluate the discriminatory power of 14 statistical features (Table~\ref{tab:60-sec_features}). Prior to the calculation, we performed label encoding on the target variable and filtered out redundant features (e.g., timestamps and distance) from the dataset to ensure the analysis only focused on the features' discriminatory power.

\begin{figure}[htbp]
  \centering
  \includegraphics[width=1\linewidth, height=5cm]{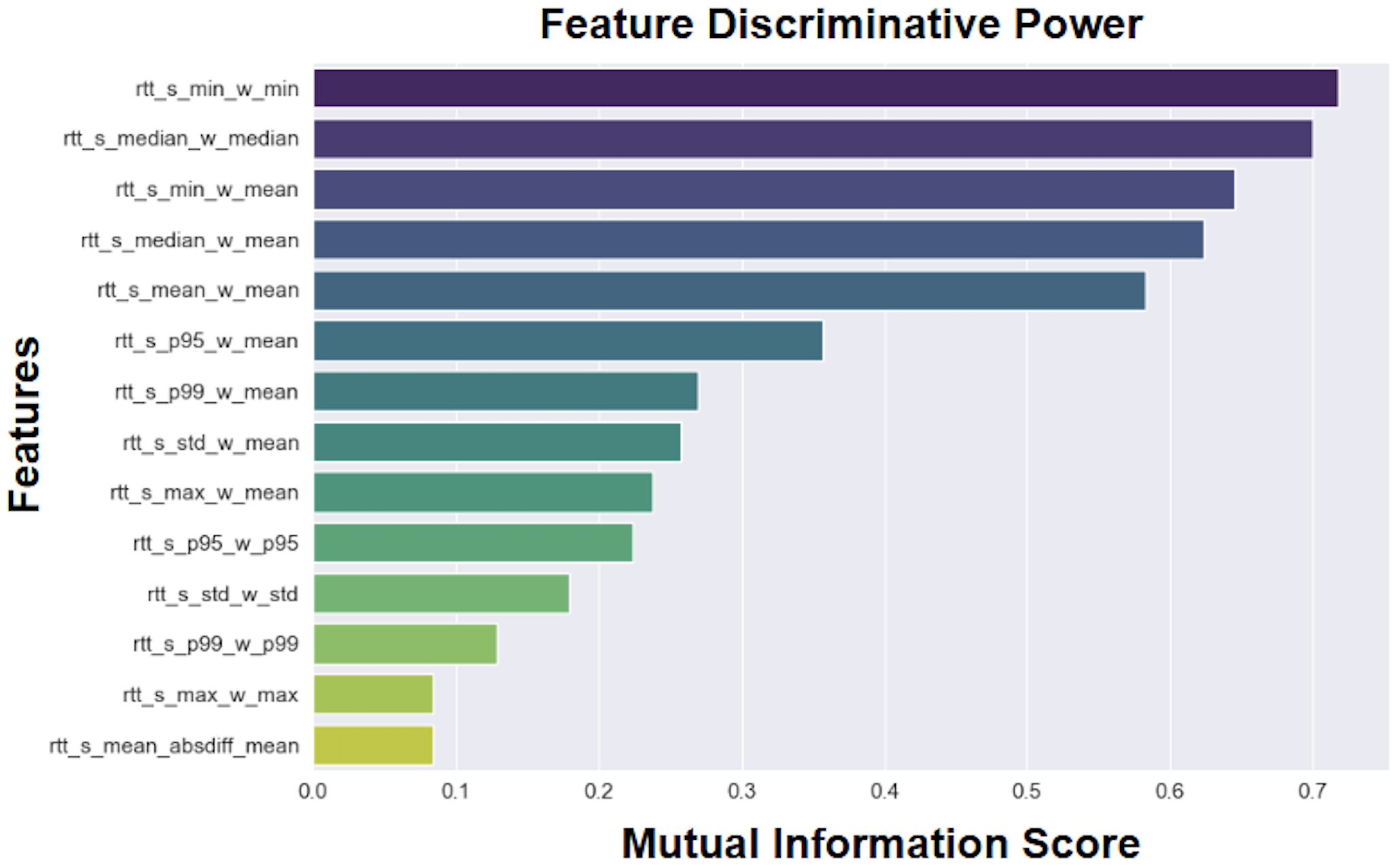}
  \caption{Mutual Information scores ranking the discriminative power of 14 secondary features across five regions.}
  \label{fig:mutual_information}
\end{figure}

The resulting MI scores are present in Fig.~\ref{fig:mutual_information}. The \texttt{rtt\_s\_min\_w\_min} feature achieves the highest score, indicating that the minimum RTT has the strongest discriminative power regarding the target region. Conversely, \texttt{rtt\_s\_mean\_absdiff\_mean} and \texttt{rtt\_s\_max\_w\_max} exhibit the lowest scores, meaning these two features possess limited relevance to the regions. In summary, the MI analysis identifies the minimum RTT as the most significant feature for target region identification among the 14 statistical features. This indicates that regional differences are most strongly reflected in the lower-bound latency behavior, which is closely related to the best achievable path condition rather than other statistical features.

\subsection{Classification Model}
We employ the Extreme Gradient Boosting (XGBoost) model \cite{chen2016xgboost} to classify RTT measurement data by region. Details regarding the dataset partition are provided in Section V-D1, while the classification accuracy and performance metrics are discussed in Section V-D2. Furthermore, to validate the MI-based analysis and quantify the discriminative power of each statistical feature, we use the XGBoost split scoring function ($L\_split$) in Section V-D3. We list four rationales for choosing XGBoost as our primary classification model:


\textbf{Feature Interpretability:} We need the model to provide which statistical features are important for classifying the regions. XGBoost's split scoring function (i.e., $\mathcal{L}_{split}$) can serve this role.
\begin{equation}
\begin{split}
\mathcal{L}_{split} &= \frac{1}{2} \biggl[ \frac{(\sum_{i \in I_L} g_i)^2}{\sum_{i \in I_L} h_i + \lambda} + \frac{(\sum_{i \in I_R} g_i)^2}{\sum_{i \in I_R} h_i + \lambda} \\\\& - \frac{(\sum_{i \in I} g_i)^2}{\sum_{i \in I} h_i + \lambda} \biggr] - \gamma
\end{split}
\end{equation}

Within the brackets, the first and second terms represent the scores of the left and right child nodes after the split, respectively. The third term is the parent node score before the split. Here, $\lambda$ and $\gamma$ are regularization parameters. By using the $\mathcal{L}_{split}$ function, we can interpret which features are important for the classification task.

\textbf{Sparsity Awareness: }Packet loss is an inherent challenge in satellite networks. When packet loss occurs, the record entity is typically displayed as none (i.e., a missing value). Unlike many traditional models that require manual imputation in the data preprocessing step. XGBoost architecture employs the sparsity aware split finding algorithm for handling the missing values in the training data.

\textbf{Computational Efficiency for Fine-grained Data: }The LENS dataset is highly fine-grained (i.e., 10 ms intervals). Even after 60-s feature extraction, the six-day training dataset still remains around 8,640 samples. XGBoost is built in a block structure for in-memory storage, where the data is presorted and stored in each block before training. This design eliminates the computation for repeated sorting during the training. For data analysis in LENS, the model's time complexity is a consideration that cannot be overlooked.

\textbf{Models Performance Comparison: }We evaluated five well-known machine learning models (i.e., XGBoost, KNN\cite{guo2003knn}, Support Vector Machines (SVM)\cite{yue2003svm}, Random Forest (RF)\cite{breiman2001random}, and Gradient Boosting Decision Trees (GBDT)\cite{ke2017lightgbm}) by using same training dataset to train these models and comparing their performance across three different test datasets. The results are presented in Table~\ref{tab:different_models_performance}. Based on the comparison, the XGBoost model's accuracy is better than the others on the first and second test datasets, while the SVM model outperformed on the third dataset.

\begin{table}[htbp]
    \centering
    \caption{Classification Accuracy of Different Models Across Different Test Datasets and Dates}
    \label{tab:different_models_performance}
    \resizebox{0.45\textwidth}{!}{
    \begin{tabular}{l|ccccc}
        \toprule
        \textbf{Test Dataset} & \textbf{XGBoost} & \textbf{KNN} & \textbf{SVM} & \textbf{RF} & \textbf{GBDT} \\
        \midrule
        Jan 24, 2026 & \textbf{0.83} & 0.73 & 0.78 & 0.80 & 0.78 \\
        Feb 04, 2026 & \textbf{0.77} & 0.68 & 0.73 & 0.75 & 0.74 \\
        Mar 10, 2026 & 0.68 & 0.66 & \textbf{0.69} & 0.67 & 0.66 \\
        \bottomrule
    \end{tabular}
    }
\end{table}

\subsubsection{Experimental Datasets}
The training dataset comprises RTT measurement data from five regions (i.e., regions in the map~\ref{fig:map}) over a six-day period (Jan 18--23, 2026). To evaluate the model’s short-term and long-term stability, the test datasets were selected from three independent test days: January 24th, 2026 (i.e., short-term), as well as February 4th and March 10th, 2026 (i.e., long-term). This design allows us to evaluate the model's performance not only on data immediately after the training period but also on data from subsequent months. Notably, the Kanazawa region's RTT measurement data in the LENS dataset has been empty since January 25, 2026. Therefore, our long-term test datasets only include four regions, excluding the Kanazawa region. Before training and evaluation, both datasets undergo the aforementioned 60-s feature extraction. Finally, the training and test datasets consist of 14 statistical features (see Table~\ref{tab:60-sec_features}) and one corresponding target label (i.e., region).

\subsubsection{Experimental Results}
Table~\ref{tab:experiment_results} shows the model’s performance on the three test datasets. The model achieves a high classification accuracy of 83\% on the short-term test dataset. Its accuracy gradually declines on long-term test datasets. This indicates that the model is highly effective for short-term classification but encounters a gradual decrease in performance over longer periods. 

\begin{figure}[htbp]
    \centering\includegraphics[width=.5\linewidth]{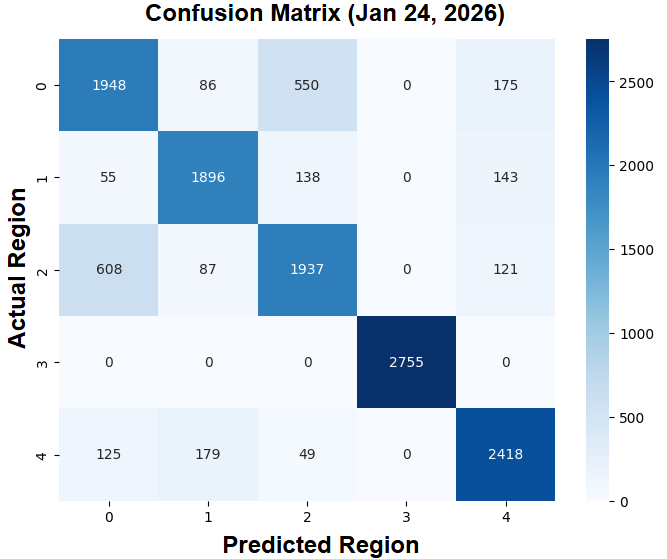} 
    \caption{Confusion matrix for short-term test dataset. The region indices: 0 (Bruhl), 1 (Kanazawa), 2 (Seattle), 3 (Ulukhaktok), 4 (Victoria).}
    \label{fig:confusion_matrix_short-term}
\end{figure}

\begin{figure}[htbp]
    \centering
    \includegraphics[width=0.48\linewidth]{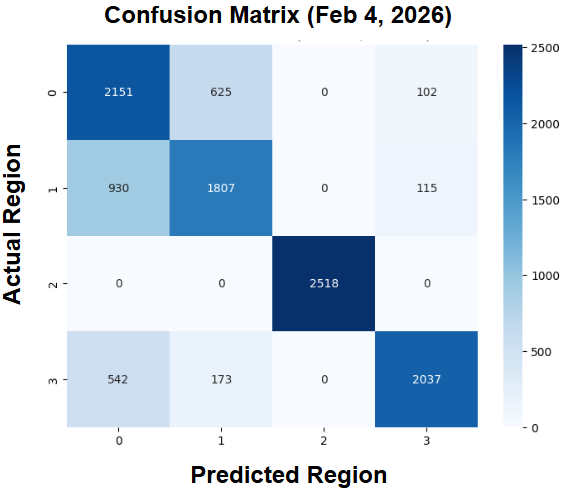} 
    \includegraphics[width=.48\linewidth]{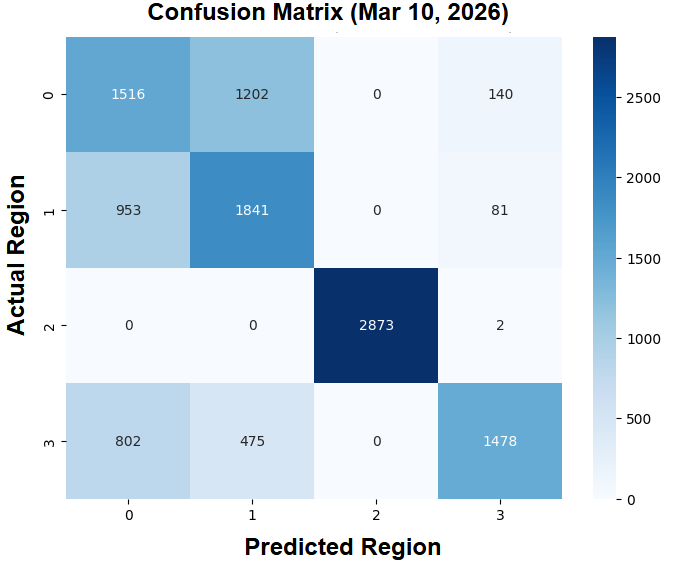}
    \caption{Confusion matrix for long-term test datasets. The region indices: 0 (Bruhl), 1 (Seattle), 2 (Ulukhaktok), 3 (Victoria).}
    \label{fig:confusion_matrix_long-term}
\end{figure}

Confusion matrices for short-term and long-term test datasets are provided in Figs.~\ref{fig:confusion_matrix_short-term} and~\ref{fig:confusion_matrix_long-term}. By analyzing the confusion matrix, we can identify which regions the model classified more accurately. We observe that region 2 (i.e., Ulukhaktok) achieves the highest scores across all three test datasets. This matched our expectations because the Cross-region Comparison section shows that the RTT statistical features for Ulukhaktok are clearly distinct from the other regions. When these differences are significant, the model can distinguish the regions more easily. In addition, from both long-term test confusion matrices (Fig.~\ref{fig:confusion_matrix_long-term}), the model encounters difficulty classifying between regions 0 and 1 (i.e., Bruhl and Seattle, respectively). This problem arises as the RTT distributions for the two regions are similar, causing the model to become confused when making predictions. 

\begin{table}[htbp]
    \centering
    \caption{Classification Performance of XGBoost Across Different Test Datasets and Test Dates}
    \label{tab:experiment_results}
    \begin{tabular}{c|c|c|c|c}
        \toprule
        \textbf{Test Dataset} & \textbf{Precision} & \textbf{Recall} & \textbf{F1-score} &\textbf{Accuracy}\\
        \hline
        Jan 24, 2026 & 0.83 &  0.83 &  0.83 &  0.83 \\
        Feb 04, 2026 & 0.80 &  0.78 &  0.78 &  0.77 \\
        Mar 10, 2026 & 0.71 &  0.68 &  0.68 &  0.68 \\
        \bottomrule
    \end{tabular}
\end{table}

\begin{figure}[htbp]
    \centering\includegraphics[width=1\linewidth, height=5cm]{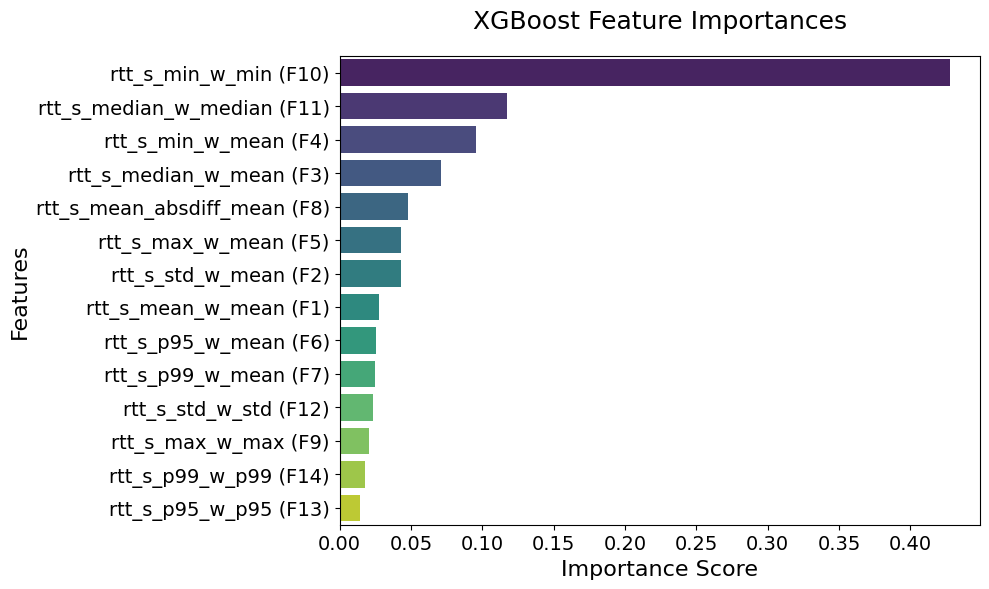}
    \caption{Trained XGBoost model feature importance scores.}
    \label{fig:model_feature_importance}
\end{figure}

\subsubsection{Features Importance}
Fig.~\ref{fig:model_feature_importance} illustrates each feature’s importance score for the trained XGBoost model. We observe that the importance score of the \textbf{rtt\_s\_min\_w\_min} feature is significantly higher than that of the other features. This result is highly consistent with the conclusion made in the Mutual Information Analysis section. Consequently, we conclude that the minimum RTT is the most significant feature for classifying RTT across different regions. 

\begin{table}[htbp]
    \centering
    \caption{Feature-Level Temporal Drift-Impact (DI) Scores Across Long-term Test Datasets}
    \label{tab:drift_impact}
    \resizebox{0.5\textwidth}{!}
    {
        \begin{tabular}{c|c|c|c}
            \toprule
            \textbf{Feature} & \textbf{XGBoost Importance} & \textbf{Feb. 04 DI}& \textbf{Mar. 10 DI}\\
            \hline
            \textbf{F10} & 0.4274   & 0.0742& 0.0296\\
            \textbf{F11} & 0.1177   & 0.0193& 0.0254\\
            \textbf{F4}  & 0.0954   & 0.0129& 0.0130\\
            \textbf{F3}  & 0.0713   & 0.0126& 0.0155\\
            \textbf{F8}  & 0.0477   & 0.0022& 0.0036\\
            \textbf{F5}  & 0.0433   & 0.0091& 0.0059\\
            \textbf{F2}  & 0.0431   & 0.0079& 0.0046\\
            \textbf{F1}  & 0.0277   & 0.0050& 0.0054\\ 
            \textbf{F6}  & 0.0257   & 0.0052& 0.0036\\
            \textbf{F7}  & 0.0246   & 0.0052& 0.0034\\
            \textbf{F12} & 0.0235   & 0.0014& 0.0019\\
            \textbf{F9}  & 0.0208   & 0.0020& 0.0021\\
            \textbf{F14} & 0.0179   & 0.0020& 0.0015\\
            \textbf{F13} & 0.0140   & 0.0021& 0.0008\\
            \bottomrule
        \end{tabular}
    }
\end{table}

\subsection{Long-term Drift-Impact Analysis}
To further investigate the degradation of long-term classification performance over time, we analyze the temporal drift of the 14 extracted RTT features. The training period is used as the reference distribution and two long-term datasets are compared against it. Specifically, for each feature and each region, we compute the standardized mean difference between the feature distribution in the training period and the test datasets. The resulting drift score measures how much a feature drifts from its training period behavior on a given test date dataset. Since not all features are equally important for classification, we further combine the drift score with XGBoost feature importance scores. The resulting drift-impact (DI) score is defined as the product of the feature-level drift and the corresponding XGBoost importance. Therefore, a larger DI score indicates that a feature is both temporally unstable and important in the region classification model. 
All DI scores are shown in Table~\ref{tab:drift_impact}. Feature F10 (i.e., \texttt{rtt\_s\_min\_w\_min}) has the highest XGBoost importance and also exhibits the largest DI scores on both Feb.04 and Mar.10 long-term test datasets. This suggests that the minimum-RTT signature is highly discriminative but also temporally unstable. These analyses determined that the learned region-level RTT signatures are not fully stationary over time. Therefore, the decreases in long-term accuracy are likely related not only to the model structure itself, but also to temporal distribution drifts in the latency features, which may be caused by changes in user demand, ground station/PoP status, network traffic, or other operational conditions.

\section{Conclusion}
This paper analyzes RTT measurements from geographical regions in the satellite Internet measurement dataset to identify regional latency patterns and evaluate whether statistical features can distinguish between regions. Results show that areas with limited infrastructure and greater distances between the dish and PoP station experience higher latency. Among the features, minimum RTT proved most effective for regional discrimination, supported by both MI analysis and XGBoost split scoring. While the model performs well in the short term, improving long-term accuracy remains a challenge. The long-term drift-impact analysis further indicates that this degradation is associated with shifts in the distribution of latency features, particularly the minimum-RTT signature, which is highly discriminative but not fully stationary over time. Future work will explore online learning approaches and additional RTT features to enhance and maintain classification performance over time.


%




\ifCLASSOPTIONcaptionsoff
  \newpage
\fi




\bibliographystyle{IEEEtran}
\bibliography{references}





%




\end{document}